\documentclass[conference]{IEEEtran}
\usepackage{cite}

\ifCLASSINFOpdf
\usepackage[pdftex]{graphicx}
\else
\fi
\usepackage[cmex10]{amsmath}
\usepackage[caption=false,font=footnotesize]{subfig}
\usepackage{url}

\usepackage{amssymb}
\usepackage{booktabs}
\usepackage{color}
\usepackage{makecell}
\usepackage{soul}
\usepackage{hyperref}

\begin{document}
	
	\definecolor{yellow}{rgb}{1.0, 1.0, 0.8}
	\definecolor{blue}{rgb}{0.0, 0.2, 0.5}
	\definecolor{dark_green}{RGB} {0, 140, 0}
	\definecolor{purple}{RGB}{128,0,128}
	\definecolor{orange}{RGB}{255,165,0}
	
	\newcommand{\JUN}[1]{\textcolor{orange}{\textbf{\small [}\colorbox{white}{\textbf{Jun:}}{\small #1}\textbf{\small ]}}}
	\newcommand{\MB}[1]{\textcolor{dark_green}{\textbf{\small [}\colorbox{white}{\textbf{Mauro:}}{\small #1}\textbf{\small ]}}}
	\newcommand{\BT}[1]{\textcolor{blue}{\textbf{\small [}\colorbox{white}{\textbf{Benedetta:}}{\small #1}\textbf{\small ]}}}
	\newcommand{\TODO}[1]{\textcolor{red}{{TODO: #1}}}
	\newcommand{\CH}[1]{\textcolor{purple}{#1}}

	\title{An Architecture for the detection of GAN-generated Flood Images with Localization Capabilities  \vspace{-0.7cm}}
	%
	\author{\IEEEauthorblockN{Jun Wang, Omran Alamayreh, Benedetta Tondi and Mauro Barni}
		\IEEEauthorblockA{Department of Information Engineering and Mathematics, University of Siena, Via Roma, 53100 Siena, Italy\\
		}
		
	}
	
	
	%


	\maketitle
	
	\begin{abstract}
		In this paper, we address a new image forensics task, namely the detection of fake flood images generated by ClimateGAN architecture. We do so by proposing a hybrid deep learning architecture including both a detection and a localization branch, the latter being devoted to the identification of the image regions manipulated by ClimateGAN. Even if our goal is the detection of fake flood images, in fact, we found that adding a localization branch helps the network to focus on the most relevant image regions with significant improvements in terms of generalization capabilities and robustness against image processing operations. The good performance of the proposed architecture is validated on two datasets of pristine flood images downloaded from the internet and three datasets of fake flood images generated by ClimateGAN starting from a large set of diverse street images.
	\end{abstract}
	

	%
	\IEEEpeerreviewmaketitle

	\section{Introduction}
	\label{sec:intro}

	\enlargethispage{\baselineskip}

	Nowadays, Generative Adversarial Networks (GANs)  can synthesise artificial images with a quality that can easily fool humans. As a consequence,  Multimedia Forensics (MF) researchers have increased their efforts to develop tools capable of detecting GAN-generated contents.
	Climate change is almost invariably considered as one of the most serious threats to humankind and one of the hardest challenges our society will have to face with in the coming years.
	The visualization of the effects of climate change can play a major role to develop new weather analysis tools and to raise public awareness. For this reason, several works have been proposed to generate weather-sensitive images using GAN networks \cite{kawakami2021semi,karacan2016learning,rothmeier2021let,lin2019weagan,schmidt2021climategan}.
	Though the purpose of these works is to promote research and raise the awareness on climate change, the risk exists that climate-sensitive synthetic images are used maliciously or exploited within organised disinformation campaigns. It is, then, of paramount importance to develop tools capable of distinguishing natural and synthetic weather images.
	

	In this paper, we present the first method for the detection of GAN-generated weather images. Specifically, we focus on the detection of flood images generated by the \textit{ClimateGAN} network, which can be used to transform general street images into images showing extreme flood conditions.
	%
	The proposed network has a hybrid architecture based on ResNet50 \cite{he2016deep}. The basic ResNet50 architecture is used to extract the features allowing the distinction between fake and natural images. Even if our goal is to distinguish natural and fake flood images, however, the features extracted by the convolutional layers of ResNet50, are also used to drive a fully convolutional architecture in charge of localizing the manipulated image region. One of the main findings of our work, in fact, is that adding a localization task on top of the feature extraction layers, has a positive effect on detection accuracy, even if localization is not very accurate. The reason for such a counterintuitive behaviour is that by asking the network to localize the manipulation, we somehow force the feature extraction layers to focus on the most significant parts of the image.
	%
	%
	The effectiveness of the proposed method was validated on the collected flood images showing a very good detection accuracy. The performance of the localisation module
	is also generally good, even if we remark that the final goal of our architecture is the detection of GAN generated images and the addition of a localisation module is mainly an expedient to force the network to look at the most significant image regions.
	Among the strengths of the proposed solution we stress that: i) the detector can be trained with a reduced number of images (400 natural and 400 fake images in our experiments); ii) the detector exhibits very good generalisation capabilities, as demonstrated by the experiments we run on natural images gathered from various sources across the Internet; iii) the proposed method shows good robustness against common image processing methods such as image resizing, addition of Gaussian noise, JPEG compression, and filtering.
	
	The rest of the paper is organized as follows. In Section \ref{sec:related}, we briefly
	overview methods for the detection of synthetic images. In Section \ref{sec:data}, we describe the datasets we have used in our experiments. The proposed forgery detector is detailed in Section \ref{sec:methodology}. The results of the experiments we carried out are reported and discussed in Section \ref{sec:experiments}. In Section \ref{sec:conclusion}, we conclude the paper with some final remarks and clues for future research.
	
	\section{Related work}
	\label{sec:related}

	Traditionally, the pipeline for the design of an effective image forgery detector consists in
	extracting a set of hand-crafted features \cite{li2020identification,mccloskey2019detecting,hu2021exposing}, and using them to detect the presence of forgery artifacts.
	For the detection of GAN-generated images and forgeries, some works have proposed to analyze the frequency artefacts introduced by GANs due to the upsampling operation applied by such networks \cite{yu2019attributing,liu2021spatial,luo2021generalizing}.
	Recently, data-driven detectors based on various CNN architectures, e.g.,\cite{rossler2019faceforensics++, chai2020makes,zhao2021multi},
	are gaining more and more popularity, due to their superior performance, at least when the test conditions match those used during training.
	%
	%
	Forgery localization has received relatively less attention, yet the possibility to localize where an image has been tampered with would be particularly useful to explain the detection results and improve the generalization capability of the detector.
	%
Most of the appraoches dealing with the localization of GAN genereated contents focus on faces, namely, deepfake.
	A first technique focusing on the localization of manipulated image areas was proposed in \cite{songsri2019complement}, by relying on paired facial landmarks information and a face image as input. Huang et al. \cite{huang2020fakelocator} proposes a \textit{FakeLocator} model to localize the texture artefacts introduced by the up-sampling steps usually applied by GAN architectures. By focusing on a specific task, i.e., face swapping, \cite{mazaheri2022detection} and \cite{nguyen2019multi} obtain tampering localization by segmenting the input images into manipulated regions, and then share the features for the detection task.
	Similarly, \textit{Face X-ray} \cite{li2020face} analyzes the artefacts introduced when two images are blended and use this information to guide the training of the detector.
Finally, Dang et al. \cite{dang2020detection} and Zhao et al. \cite{zhao2021multi} use an attention mechanism to improve the detection accuracy and to identify the image region that mostly contributes to the detection.
	In our work, we propose a hybrid framework, wherein localization is used to aid the detection of GAN manipulated flood images
	To the best of our knowledge, this is the first work focusing on the detection of GAN manipulated flood images.
	
	\enlargethispage{\baselineskip}
	
	\section{Dataset construction}
	\label{sec:data}

	
	\subsection{Natural flood images}
	
	With regard to natural flood images, we relied on two public datasets released recently.
	%
	The \textbf{Roadway Flooding Image (RWFI)} dataset \cite{sazara2019detecting} consists of 441 images with different scenes from urban, suburban and natural settings.
	The images have all the same size equal to 385$\times$512. Mask images of flooded areas are also available. We used this dataset for training.
	The \textbf{WSOC Flood Image} dataset \cite{zaffaroni2020water} contains 490 real flood images gathered from Twitter and online news using "floods" as the keyword for the search query. 439 of such images are also accompanied by a binary mask image identifying the flooded areas. 
	We used these 439 flood images to test the generalization capability of the proposed detector.
	
	\subsection{ClimateGAN Images}
	\label{sec:climate}

\begin{figure}
		\centering
		\includegraphics[width=0.95\columnwidth]{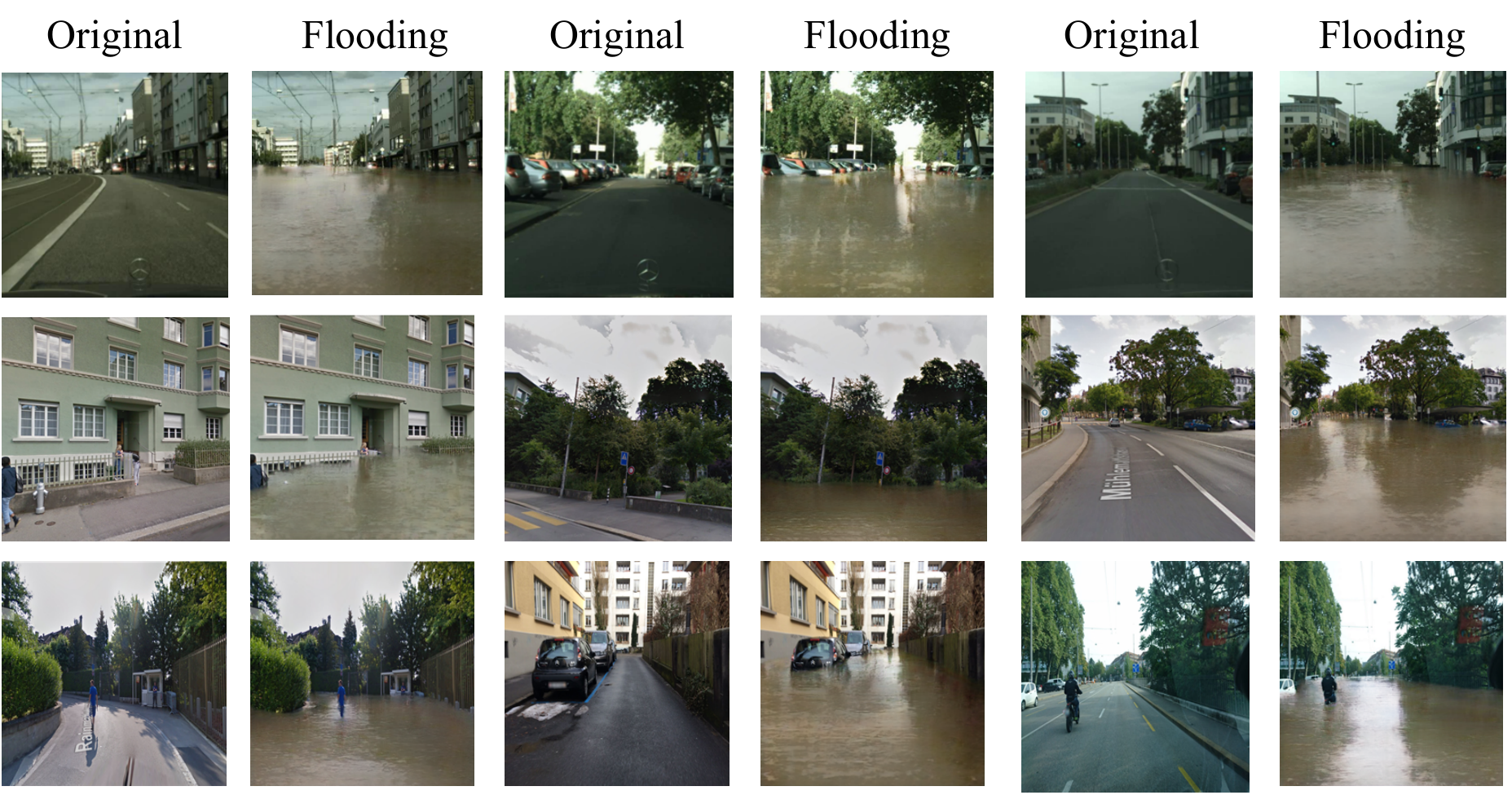}
		\vspace{-0.3cm}
		\caption{GAN generated flood images. Top: StreetG; Middle: WebG132; Bottom: WebG504.}
		\label{fig:gan}
	\end{figure}

	We focused on the detection of synthetic flood images generated by  the \textit{ClimateGAN} network described in \cite{schmidt2021climategan}.
	\textit{ClimateGAN} is an architecture for image-to-image translation whose goal is to produce images with extreme flooding conditions from street images.
	%
	%
	By using the pre-trained \textit{ClimateGAN} model released by the authors of \cite{schmidt2021climategan}\footnote{\url{https://github.com/cc-ai/climategan}}, we built three synthetic flood image datasets
	by using diverse street images. Specifically, we first created 3,900 GAN flood images (hereafter referred to as \textbf{StreetG} dataset), starting from the Cityscapes \cite{cordts2016cityscapes} and Kitti \cite{geiger2013vision} video datasets described in \cite{Alamayreh2021gan}.
	For each video, we took a frame every 60 frames and used the central cropped area of size 512 $\times$ 512 as input of the \textit{ClimateGAN}. We also built two additional small \textit{ClimateGAN} datasets, namely the \textbf{WebG132} and \textbf{WebG504} datasets, to validate the generalization capability of the proposed detector. The street view images used to build these datasets were collected from the internet by using the Mapillary tool (\url{https://www.mapillary.com}). More specifically, WebG132 contains 132 \textit{ClimateGAN} flood images built as  \textbf{StreetG}. WebG504 is an extended version of WebG132 built by resizing the input images rather than cropping. WebG504 consists of 504 GAN flood images.
	%
	Some examples of the fake images contained in the datasets are given in Fig. \ref{fig:gan}. The ClimateGAN image datasets split in training and testing are available at the link \url{https://drive.google.com/drive/folders/12Qd1-T8J0jCwYYqiOoARsahyElB--Jai?usp=sharing}.
	\begin{figure}[t]
		\centering
		\includegraphics[width=1\linewidth]{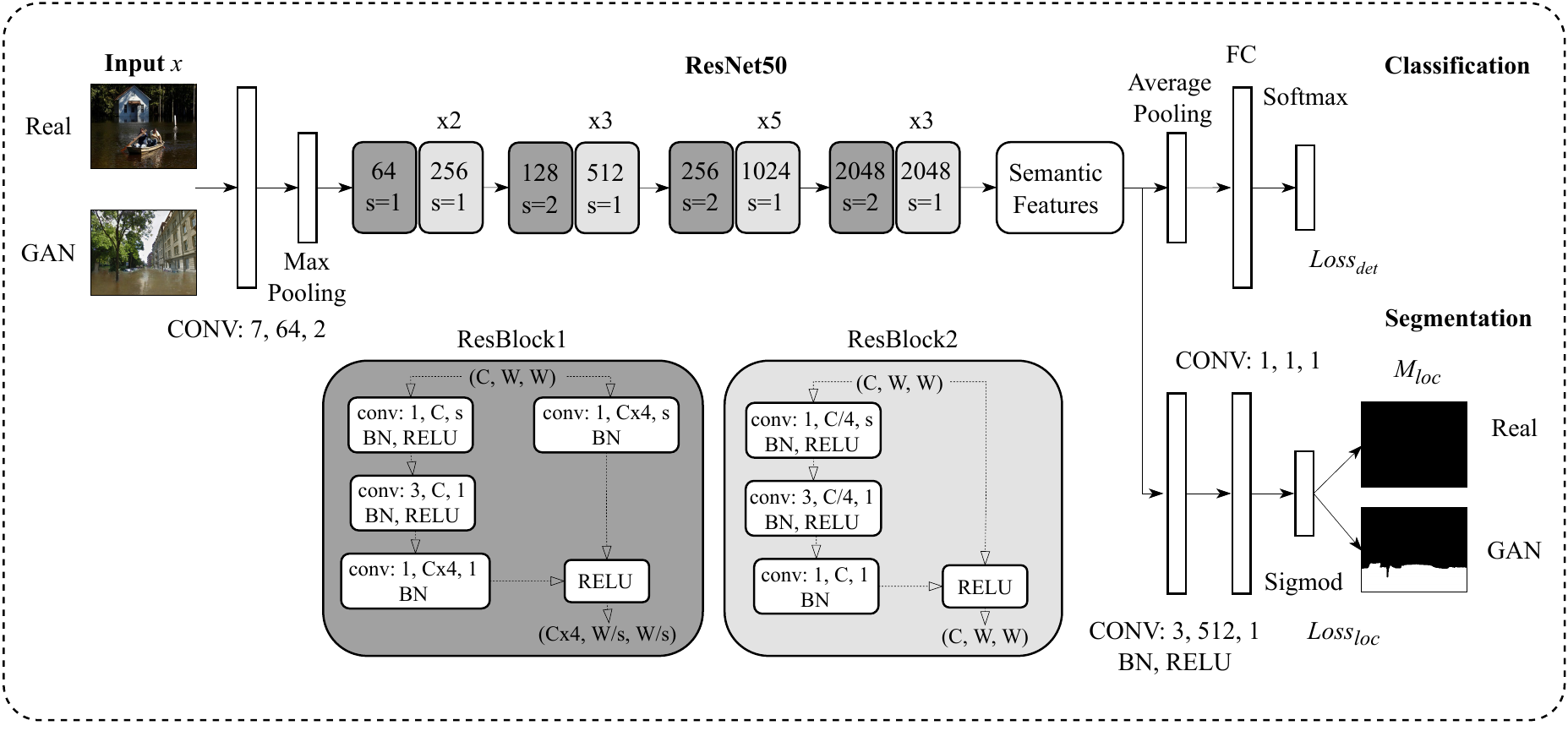}
		\vspace{-0.3cm}
		\caption{Overview of the proposed detection/localization architecture.}
		\label{fig:framework}
		\vspace{-0.3cm}
	\end{figure}
	
	\enlargethispage{\baselineskip}

	\section{The Proposed Detector}
	\label{sec:methodology}
	
	In this section, we present our framework for the detection of GAN-manipulated flood images and the localization of the manipulated flooded areas.
	%
	Given a flood image $x$, with $x \in \mathbb{R}^{H \times W \times 3}$, the goal of our system is to associate to $x$ a label $y$ and a mask $M$, in such a way that $y =1$, if $x$ is a manipulated image and $0$ otherwise, and the mask $M$ indicates the pixels where the image has been manipulated (pixels for which $M = 1$ indicate the manipulated areas).
	The main intuition behind our work is that asking the network to localize the manipulated areas also helps the detection process, since, during training, we can instruct the network to focus on the image areas that have been most likely been manipulated, namely the water regions.
	The overall architecture of the proposed method is shown in Fig. \ref{fig:framework}.
	
	To start with, given an image $x$, we pass it through a backbone ResNet50 network to extract semantic features $SF$.
	%
	The features extracted in this way are exploited for the detection and localization tasks. Specifically, for the localization task, we consider a fully convolutional network (FCN) \cite{long2015fully},
	where the localization map $M_{loc}$ is obtained by using two convolutional layers and a sigmoid function layer. At the same time, the semantic features $SF$ obtained from the last convolutional layer of ResNet50 are sent to an AvgPool layer and a Fully Connected (FC) layer to distinguish between GAN and real images.
	In order to train the proposed architecture, we rely on two loss functions,
	one associated with the detection task and one with the localization task. The detection loss is defined as:
	\begin{equation}
		L_{det}=\frac{1}{N}\sum_{i}Y_{i} \log(y_i) + (1-Y_{i})\log(1-y_i)
	\end{equation}
	where the sum is extended to all the images in the training set, $y_i$ is the output of the detector in correspondence to the input image $x_i$, and $Y_i$ is the corresponding ground truth,  which determines whether or not an image is manipulated.
	In a similar way, the localization loss is defined as
	%
	\begin{equation}
		L_{loc}=\frac{1}{S}\sum_{i}^{N}\sum_{j}^{H\times W}G_{ij} \log(M_{ij}) + (1-G_{ij})\log(1-M_{ij})
	\end{equation}
	%
	with $S = N\times H\times W$, and where $M_i \in \mathbb{R}^{H \times W}$ is the estimated manipulation map for the test image $i$, $M_{ij}$ the value assumed by the map in correspondence of the $j$-th pixel, and $G_{ij}$ the corresponding ground-truth.
	The total loss used to train the model, then,  is the weighted sum of the two losses above, namely:
	%
	$L_{tot} = \lambda _{det} \cdot L_{det} + \lambda_{loc} \cdot L_{loc}$,
	%
	where $\lambda_{det}$ and $\lambda_{loc}$ must be set in such a way to balance the importance of the two loss terms.
	
	
	\section{Experiments}
	\label{sec:experiments}
	\subsection{Setting}
	\textbf{Baseline methods.}
	To prove the effectiveness of the proposed method, we selected two popular deep neural networks, namely Xception and ResNet50, which have been proven to be effective for several forgery detection applications (see, for instance,  \cite{Alamayreh2021gan} and \cite{gragnaniello2021gan}). We considered various training strategies. We trained
	an Xception and ResNet50 model as standard binary detectors, to decide if the input image has been generated by \textit{ClimateGAN} or not. In the ResNet50 case, the architecture coincides with the classification branch of the proposed hybrid architecture. Besides, we also trained four models by using an additional mask image as an input, to guide the attention of the network toward water regions. In detail,  two models (referred to with the label MUL) take as input the image $x$ multiplied by a binary mask indicating the water regions. The other two models (referred to with the label CAT), take as input a 4-channel image, with the fourth band containing a binary mask highlighting the water areas.
	The performance obtained by the above networks are compared to the detection performance of our hybrid architecture, hereafter referred to as ResNet50\_hyb.
	It is worth stressing that, in contrast to the networks taking as an additional input the mask with the water regions, our method does not need such information. In fact, we force the network to look at the water regions indirectly, by asking the network to localise the manipulated areas.
	
	\textbf{Implementation details}.
	All the models were trained with the same configuration from the scratch for 30 epochs.
	For the optimization, we used the Adam optimizer with a learning rate of 0.0001, a mini-batch size of 16. The input images were resized to 224$\times$224$\times$3 and normalized with mean [0.485, 0.456, 0.406] and variance [0.229, 0.224, 0.225]. The training images were augmented by [saturation, brightness, contrast] adjustment with factors equal to [0.05, 0.05, 0.05].
	We trained the networks by using the RWFI dataset for real images and an equal number of GAN-generated flood images from the StreetG dataset. The training datasets were split into training and validation subsets with percentages 80\% and 20\% respectively. All the other datasets were used for testing.
	For the proposed hybrid network, we experimentally set $\lambda_{det} = 0.4$ (and $\lambda_{loc} = 0.6$), that is the setup that gave the best results.
	To evaluate the detection effectiveness, we consider the true positive (TPR) and true negative rates (TNR) of the decision.
	We also calculate the area under the ROC curve (AUC) performance by pairing each fake image dataset with the real image dataset. To evaluate the localization performance, we use the balanced Pixel Accuracy (bPA) and Intersection over Union (IoU), which for a binary classifier are defined as:
	\begin{equation}
		\text{bPA}=\frac{1}{2N_t}\sum_{t=1}^{N_t}\sum_{i=0}^{1} \frac{p_{ii}}{\sum_{j=0}^{1}p_{ij}},
	\end{equation}
	and
	\begin{equation}
		IoU = \frac{1}{N_t}\sum_{t=1}^{N_t}\frac{\left|\left\{ M_t \equiv 1 \right\} \cap \left\{ G_t \equiv 1 \right\} \right|}{\left|\left\{ M_t \equiv 1 \right\} \cup \left\{ G_t \equiv 1 \right\} \right|},
	\end{equation}
	where $p_{ij}$ indicates the number of pixels in class $i$ classified as class $j$, $\{M_t \equiv 1\}$ is the set of pixels of value 1 in the $M_t$ mask images,
	and $N_t$ indicates the number of test images.


	\subsection{Results}
	\begin{table}
		\centering
		\footnotesize
		\renewcommand\arraystretch{0.8}
		\setlength{\tabcolsep}{0.55mm}
		\caption{Results (\%)  for different datasets.}
		\vspace{-0.2cm}
		\begin{tabular}{@{}lccccccc@{}}
			\toprule
			& WSOC & \multicolumn{2}{c}{StreetG} & \multicolumn{2}{c}{WebG132} & \multicolumn{2}{c}{WebG504} \\
			\midrule
			Methods & TNR\% & TPR\% & AUC\% & TPR\% & AUC\% & TPR\% & AUC\% \\
			\midrule
			Xception & 96.8 & 98.0 & 99.5 & 37.9 & 84.3 & 57.7 & 91.8 \\
			ResNet50 & 95.0 & 98.6 & 99.4& 47.0 & 84.0 & 64.9 & 91.5 \\
			Xception+M (CAT) & 94.5 & 86.4 & 96.6 & 41.7 & 75.5 & 65.9 & 89.2\\
			Xception+M (MUL) & 96.8 & 97.5 & 99.5 & 65.1& 95.9 & 73.4 & 96.9\\
			ResNet50+M (CAT) & 92.0 & 99.6 & 98.7 & 47.0 & 83.0 & 60.5 & 88.6\\
			ResNet50+M (MUL) & 95.2 & 98.3 & 99.3 & 70.5 & 96.1 & 79.2 & 96.7\\
			ResNet50\_hyb & \textbf{98.0} & \textbf{100} & \textbf{100} & \textbf{93.4} & \textbf{99.0} & \textbf{95.4} & \textbf{98.9}\\
			\midrule
			Methods & bPA & bPA & IoU & bPA & IoU & bPA & IoU \\
			\midrule
			ResNet50\_hyb & 98.6 & 96.1 & 92.0 & 84.4 & 62.5 & 87.9& 71.2\\
			\bottomrule
		\end{tabular}
		\label{tab:1}
		\vspace{-0.1cm}
	\end{table}
	\begin{figure}
		\centering
		\includegraphics[width=0.9\columnwidth]{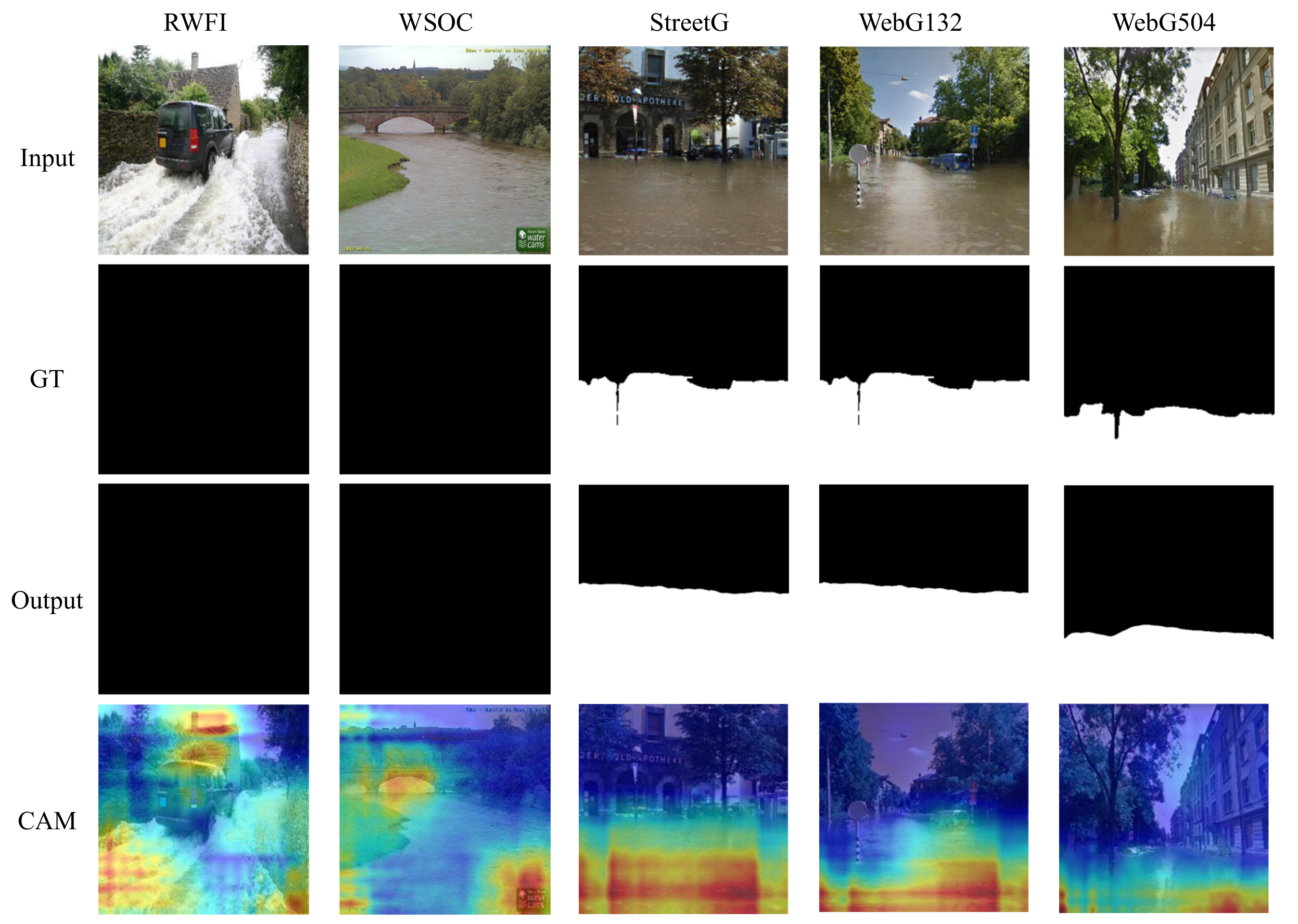}
		\caption{Visualization of output masks and CAM maps for ResNet50\_hyb.}
		\label{fig:mask}
		\vspace{-0.2cm}
	\end{figure}
	
	\enlargethispage{\baselineskip}
	
	\begin{figure*}[h!]
		\centering
		\includegraphics[width=0.495\linewidth]{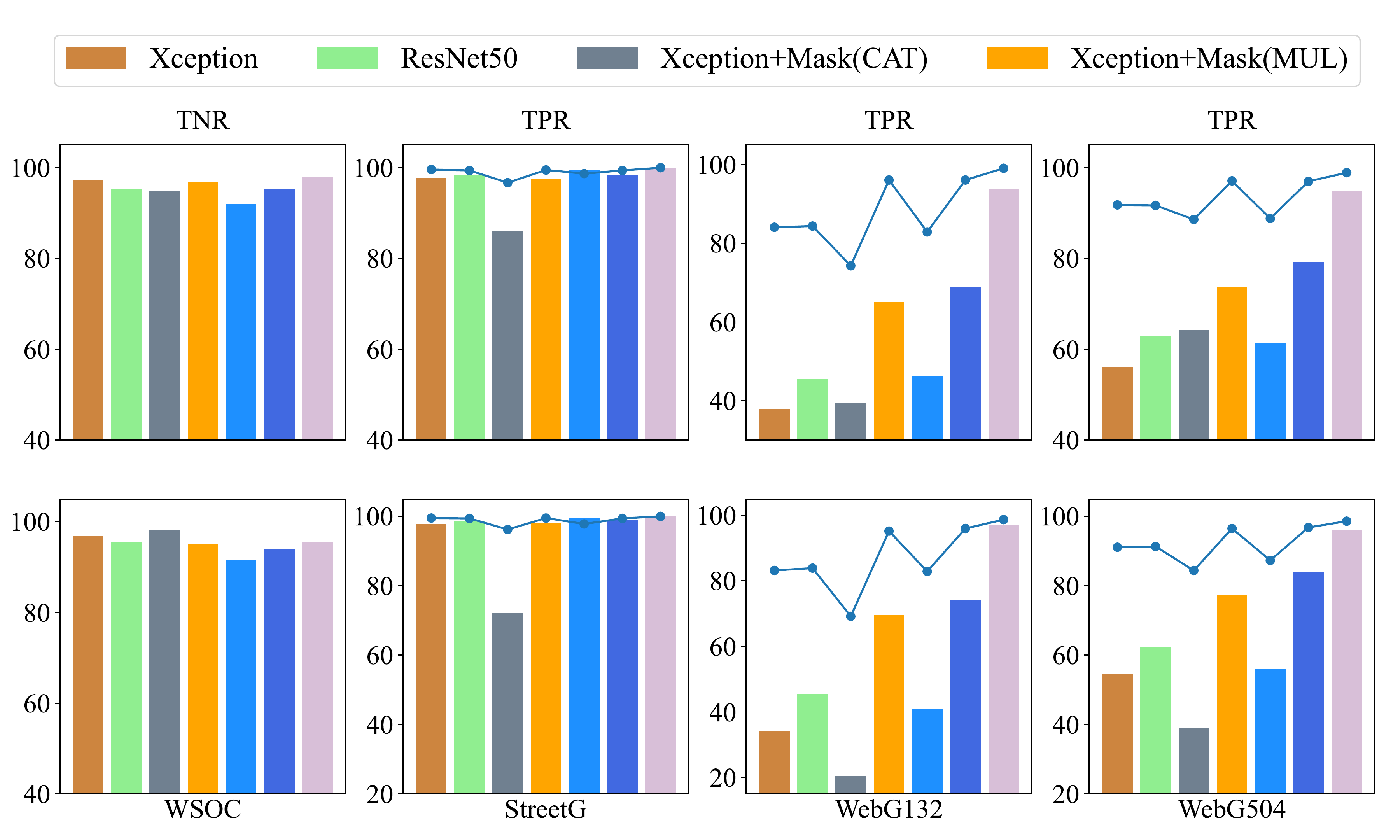}
		\includegraphics[width=0.495\linewidth]{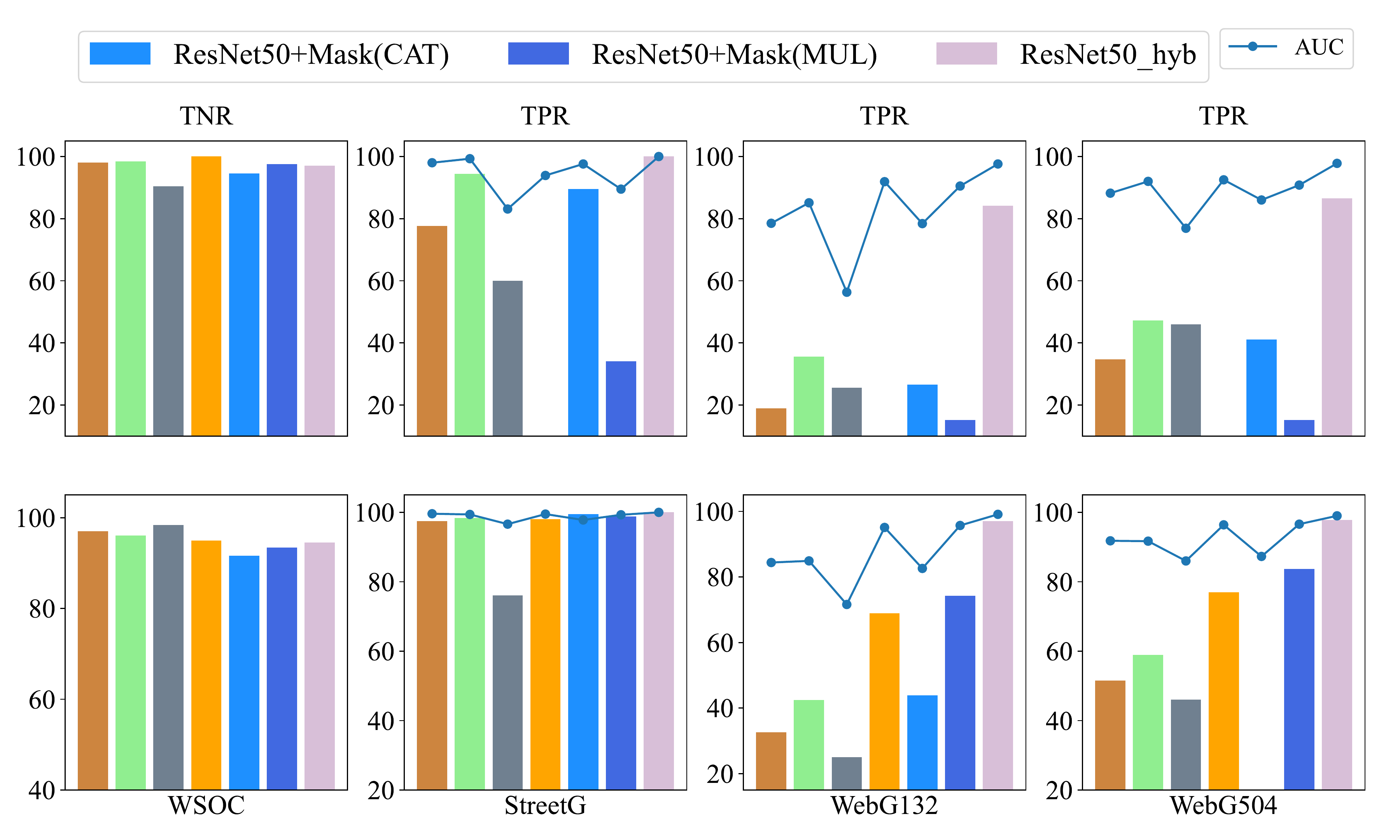}
		\caption{Robustness results (\%) in terms of TNR, TPR, and AUC in the presence of JPEG compression with a quality factor of 50 (top, 1-4), resizing with factor 0.5 (bottom, 1-4), Median filtering, window size = 3$\times$3 (top, 5-8) and Gaussian blur, window right size = 3$\times$3(bottom, 5-8).}
		\label{fig:robust2}
		\vspace{-0.2cm}
	\end{figure*}

	Table \ref{tab:1} shows the results on the four test datasets. The baseline methods achieve good classification performance on the StreetG dataset, with a TPR around 98\% and an AUC equal to 99\%. However, the performance dropped on the datasets that were not used for training. The performance of ResNet50 and Xception are the worst, likely due to the small number of images available for training.
	The way the water mask images are used during training also has a significant impact on the performance. By looking at the results in rows 3 to 6, multiplying the mask and the input image results in better performance on WebG132 and WebG504 datasets, than simply concatenating the mask to the image bands.
	%
	The hybrid detection/localization method proposed in this paper (ResNet50\_hyb) achieves the best performance on all the datasets, with accuracies always well above 90\% and often close to 100\%. In particular, the proposed method shows a good generalization capability, achieving very good performance on WebG132 and WebG504 datasets, with 93.4\% and 95.4\% TPR values, and 99.0\% and 98.9\% AUCs. This is a remarkable result, given that such good performance is obtained without the additional information provided by the water mask.
	The superior performance provided by the hybrid model is even more interesting if we consider the localization accuracy. By looking at the bPA and IoU values reported in
	the last row of Table \ref{tab:1}, we can see that the proposed network provides good results on WSOC and StreetG, even if we can notice a performance drop on WebG132 and WebG504 datasets. The remarkable conclusion we can draw is that asking the network to localize the tampered regions helps to distinguish fake images from real ones, even when localization does not work very well,
	since it forces the detection branch to focus on the manipulated area.
	Such intuition is confirmed by the CAM maps \cite{selvaraju2017grad}, computed with regard to the detection task, shown in the last row of Fig. \ref{fig:mask}. 

	Table \ref{tab:3} and Figure \ref{fig:robust2} show the robustness of the methods against various image processing operators, including JPEG compression with quality factor 50, image downsampling with resizing 0.5, Median filtering (3$\times$3), Gaussian blur (3$\times$3), and Gaussian noise addition with zero mean and variance equal to 0.003\footnote{In all our experiments images take values in the [0,1] range.}.
	%
	Overall, all the models are robust to the image processing methods used in our tests, with the exception of Gaussian noise addition (see Table \ref{tab:3}). In the Gaussian noise case, in fact, the performance of the Xception+mask (MUL) model dropped from 97.5\%, 65.1\% and 73.4\% to 1\%, 0\%, and 0\%.
	ResNet50+mask (MUL) also suffers the addition of Gaussian noise. The bad behaviour of the models adopting a multiplicative approach to include the information provided by the water mask indicates that, in the absence of information about the non-tampered regions, the impact of noise is amplified, since the texture information only is not enough to detect the presence of tampering. As further evidence of this intuition, we observe that Xception and ResNet50 show stronger robustness. The positive effect of considering the entire image rather than the water areas only is evident also with regard to the proposed hybrid method, which, as it can be seen from Table \ref{tab:3}, retains good performance also in the presence of noise addition.
	%
	With regard to the other attacks (see Figure 
	\ref{fig:robust2})
	the proposed method always achieves the best results with a minor performance drop on the WSOC real dataset only.
	
	\begin{table}
		\centering
		\footnotesize
		\renewcommand\arraystretch{0.8}
		\setlength{\tabcolsep}{0.6mm}
		\caption{Results (\%) in the presence of Gaussian noise.}
		\vspace{-0.2cm}
		\begin{tabular}{@{}lccccccc@{}}
			\toprule
			& WSOC & \multicolumn{2}{c}{StreetG} & \multicolumn{2}{c}{WebG132} & \multicolumn{2}{c}{WebG504} \\
			\midrule
			Methods & TNR\% & TPR\% & AUC\% &TPR\% & AUC\% & TPR\% & AUC\% \\
			\midrule
			Xception & 98.0 & 77.7 & 98.0 & 18.9 & 78.5 & 34.7 & 88.2 \\
			ResNet50 & 98.4 & 94.4 & 99.3 & 35.6 & 85.1 & 47.2 & 92.0 \\
			Xception + M (CAT) & 90.4 & 60.0 & 83.1 & 25.6 & 56.3 & 46.0 & 76.9\\
			Xception + M (MUL) & \textbf{100} & 1.0 & 93.9 & 0 & 91.9 & 0 & 92.5\\
			ResNet50 + M (CAT) & 94.5 & 89.5 & 97.6 & 26.5 & 78.4 & 41.1 & 86.0\\
			ResNet50 + M (MUL) & 97.5 & 34.1 & 89.5 & 15.1 & 90.5 & 15.1 & 90.8\\
			ResNet50\_hyb & 97.0 & \textbf{100} & \textbf{100} & \textbf{84.1} & \textbf{97.6} & \textbf{86.5} & \textbf{97.8}\\
			\bottomrule
		\end{tabular}
		\label{tab:3}
		\vspace{-0.5cm}
	\end{table}
	
	\section{Conclusion}
	\label{sec:conclusion}
	
	In this paper, we have presented a ResNet50-based hybrid network for the detection of GAN-generated flood images. In particular, we exploited localization during training to help the network to focus on the most significant parts of the analyzed image, with a positive impact on the detection performance. To train and test the proposed network, we have gathered more than 4,000 fake flood images generated by \textit{ClimateGAN} and 880 genuine flood images. The experiments we carried out show the excellent performance of our network, its good generalization capabilities, and its robustness against various image processing operations.
	As a future work, we are planning to consider different generative models. Being our approach general, we also plan to extend it to the detection of different GAN-generated climate disasters and forgeries.
	Meanwhile, it would be interesting to combine this work with geolocalization, that is, given a flood image, first localize the geographical location the image is referring to, and then verify the plausibility that a flood actually occurred.
	\section*{Acknowledgment}
	
	This work has been partially supported by the China Scholarship Council(CSC) , file No. 202008370186, by the
	PREMIER project under contract PRIN 2017 2017Z595XS-
	001, funded by the Italian Ministry of University and Research,
	and by the Defense Advanced Research Projects Agency (DARPA) and the
	Air Force Research Laboratory (AFRL) under agreement number FA8750-20-2-1004. The U.S. Government is authorized to
	reproduce and distribute reprints for Governmental purposes
	notwithstanding any copyright notation thereon.

	
	
	\bibliographystyle{IEEEtran}
	\bibliography{refs}
\end{document}